%% file: acl_latex.tex
\documentclass[11pt]{article}

\usepackage[preprint]{acl}

\usepackage{times}
\usepackage{latexsym}
\usepackage{amsmath}

\usepackage[T1]{fontenc}

\usepackage[utf8]{inputenc}

\usepackage{microtype}

\usepackage{inconsolata}

\usepackage{graphicx}

\usepackage{comment}
\usepackage{dirtytalk}
\usepackage{multirow}
\usepackage{array}
\usepackage{amsfonts}
\usepackage{threeparttablex}
\usepackage{fancyvrb}
\usepackage{fvextra}

\newcommand{\pb}[1]{\vspace{0.75ex}\noindent{\bf \em #1}\hspace*{.3em}}

\title{Small Language Model Helps Resolve Semantic Ambiguity of LLM Prompt}

\author{
 \textbf{Zhenzhen Huang\textsuperscript{1}},
 \textbf{Chaoning Zhang\textsuperscript{1,*}},
 \textbf{Fachrina Dewi Puspitasari\textsuperscript{1}},
 \textbf{Jiaquan Zhang\textsuperscript{1}},\\
 \textbf{Yitian Zhou\textsuperscript{1}},
 \textbf{Shuxu Chen\textsuperscript{2}},
 \textbf{Yang Yang\textsuperscript{1}}
\\
\\
 \textsuperscript{1}School of Computer Science and Engineering, University of Electronic Science and Technology of China,\\
 \textsuperscript{2}Department of Electronic Engineering, Kyung Hee University\\
 \small{
   \textbf{*Correspondence:} \href{mailto:chaoningzhang1990@gmail.com}{chaoningzhang1990@gmail.com}
 }
}

\begin{document}
\maketitle
\begin{abstract}
\input{sections/abstract}
\end{abstract}

\section{Introduction}
\input{sections/introduction}

\section{Related Work}
\input{sections/related}

\section{Method}
\input{sections/method}

\section{Analysis on Attention Distribution}
\input{sections/theory}

\section{Results}
\input{sections/results}

\section{Conclusion}
\input{sections/conclusion}

\section{Limitation}
\input{sections/discussion}

\bibliography{custom}

\clearpage

\appendix

\section{Appendix}\label{sec:appendix}
\setcounter{table}{0}
\renewcommand{\thetable}{A\arabic{table}}
\input{sections/appendix}

\end{document}

%% file: sections/abstract.tex
Large language models (LLMs) are increasingly utilized in various complex reasoning tasks due to their excellent instruction following capability.
However, the model's performance is highly dependent on the open-ended characteristics of the users' input prompt.
Natural prompts often do not follow proper syntactic rules, which creates ambiguous queries that yield multiple interpretations.
Such ambiguous prompts confuse the model in choosing the correct reasoning paths to answer questions.
Prior works address this challenge by applying query editing during the LLM inference process without explicitly solving the root cause of the ambiguity.
To address this limitation, we propose a pre-inference prompt optimization mechanism via explicit prompt disambiguation.
Particularly, we identify semantic risks in the prompt, check their multi-perspective consistency, and resolve any semantic conflicts that arise.
Finally, we organize the resolved ambiguities in a logically structured manner as a clean input to the LLM.
By explicitly resolving semantic ambiguity, our method can produce a more focused attention distribution to the semantically essential tokens.
We also leverage small language models (SLMs) as the main executor of prompt disambiguation to benefit from their efficient computation.
Through comprehensive experiments on multiple benchmarks, we demonstrate that our method improves reasoning performance by 2.5 points at a cost of only \$0.02.
Our study promotes explicit prompt disambiguation as an effective prompt optimization method without disturbing the internal mechanism of LLM inference.

%% file: sections/introduction.tex
Large language models (LLMs) have recently demonstrated remarkable capabilities in following instructions and performing complex reasoning across a wide range of tasks, including question answering, planning, and decision making~\cite{wei2022chain,zhou2026look}.
As these models are increasingly deployed in open-ended and interactive settings, user prompts have become the primary interface through which intent, constraints, and assumptions are supplied to the model.
However, real-world prompts are often underspecified, ambiguous, and internally inconsistent.
This condition poses a fundamental challenge to reliable reasoning~\cite{min2020ambigqa, ji2023survey}.

Semantic ambiguity in user prompts can significantly disrupt LLM reasoning by introducing multiple competing interpretations that the model must implicitly resolve during generation~\cite{min2020ambigqa, si2022prompting}.
Such ambiguity often leads to attention dispersion, unstable reasoning trajectories, and high variance across stochastic samples~\cite{dziri2022origin, ribeiro2020beyond}.
These particularly happen in tasks that require reference resolution, implicit assumption handling, or multi-step logical consistency reasoning~\cite{levesque2012winograd, zhou2020evaluating}.
As LLMs are trained to infer meaning rather than explicitly validate the prompt, ambiguous inputs may follow spurious reasoning paths that are logically plausible but globally inconsistent~\cite{turpin2023language, creswell2022faithful}.
Recent studies on multimodal reasoning further suggest that such ambiguity can be amplified when aligning heterogeneous modalities, where inconsistent semantic grounding leads to compounding reasoning errors across modalities~\cite{zheng2026LLaVA}.
As a result, even strong instruction-following models can produce brittle outputs when faced with ambiguous inputs, which limits the model's reliability in reasoning-intensive tasks.

Existing approaches largely address the challenge of prompt ambiguity through prompt engineering and prompt optimization.
These include chain-of-thought prompting~\cite{wei2022chain}, self-consistency~\cite{wang2022self}, 
semantic parsing~\cite{stengel2023zero}, and agent- or gradient-based prompt search~\cite{yuksekgonul2024textgrad,wang2026efficient,zhang2026learning,zhang2026lightweight}.
While these techniques can improve average task performance, they fundamentally rely on the LLM itself to internally resolve semantic ambiguity during inference.
As a result, they do not explicitly detect semantic risks, reconcile conflicting interpretations, or enforce input-level consistency.
This reliance on the LLM limits the methods' robustness under highly ambiguous inputs, which leads to substantial output variance and increased computational cost due to repeated large-model inference~\cite{wang2026stream,zhang2026lightweight}.
Consequently, prior methods treat ambiguity as a downstream reasoning issue~\cite{tang2025clarifying} or interaction mechanism~\cite{pilault2023interactive} rather than an upstream input-quality problem, thus leaving a gap for approaches that proactively resolve semantic uncertainty before execution.

To address this gap, we propose an upstream prompt optimization mechanism that resolves ambiguities in users' input prompts. 
We refer to our method as DisambiguSLM.
We utilize SLM for its good instruction following capability at a much lower cost than the LLM as the main model~\cite{srivastava2025thinkslm}.
SLM acts as an evaluator that examines the semantic fitness and clarity of the prompt via three-stage refinement.
Particularly, it first identifies the semantic risk points present in the prompt.
The model then passes each risk point to the next stage, where the risk goes through an interpretation consistency check and ambiguity resolution.
SLM then aggregates the resolved risk points in a logically structured manner and utilizes them as the supplementary representation to the original input prompt, as the semantically clear input to the LLM.

By passing the input prompt to DisambiguSLM and sending the output to the LLM, we can increase the LLM's reasoning performance by about 8 points of accuracy against na\"ive prompting and 3 points of accuracy against state-of-the-art prompt optimization methods.
Our method presents the largest gain at ambiguous reasoning tasks, exhibiting the benefit of DisambiguSLM in resolving semantic risks in the prompt.
We list the contribution of our study as follows:
\begin{itemize}
    \item To the best of our knowledge, we are the first to introduce a pre-inference prompt optimization method via explicit semantic disambiguation of the prompt.
    \item We propose DisambiguSLM, a prompt optimization method that resolves the semantic ambiguity in the input prompt through the identification and resolution of risk points before going through LLM inference.
    \item Through extensive evaluation on various reasoning benchmarks, we demonstrate the effectiveness of our method against strong state-of-the-art methods in improving the reasoning process of LLM.
\end{itemize}

%% file: sections/related.tex
\pb{LLM Reasoning Paths and Stability.}
Recent studies have investigated how LLMs form and follow reasoning paths during generation, revealing that attention allocation and intermediate token trajectories play a crucial role in reasoning success~\cite{wiegreffe2019attention, clark2019does,wang2025think}.
Works on reasoning, faithfulness, and robustness show that LLMs often generate answers through inconsistent or spurious reasoning paths, especially under ambiguous inputs~\cite{creswell2022faithful, turpin2023language,li2026right}.
Multi-path and tree-based reasoning frameworks, such as ReAct~\cite{yao2022react}, Tree-of-Thought~\cite{yao2023tree}, and Graph-of-Thought~\cite{besta2024graph}, explicitly explore multiple reasoning branches to improve reliability.
While effective, these methods incur substantial inference overhead and still assume that the input semantics are fixed and well-formed~\cite{zhang2026tdarc}.
Consequently, they address ambiguity implicitly through exploration rather than explicitly stabilizing the semantic space beforehand.

\pb{Prompt Optimization.}
Prompt optimization has been widely studied as a means of improving LLM performance without model retraining.
Early works, such as Chain-of-Thought~\cite{wei2022chain} prompting, demonstrate that explicitly eliciting intermediate reasoning steps can significantly improve performance on complex reasoning tasks~\cite{kojima2022large}.
Subsequent approaches extend this idea through self-consistent sampling~\cite{wang2022self}, prompt ensembling~\cite{hu2025dipper, zhang2024prefer}, and iterative refinement~\cite{madaan2023self, han2024pive}.
More automated methods have been proposed, including automatic prompt engineering (APE)~\cite{zhou2022large}, optimization via reinforcement or gradient-based signals~\cite{yuksekgonul2024textgrad}, and agent-driven prompt search frameworks such as PromptAgent~\cite{wang2023promptagent} and OPRO~\cite{yang2023large}.
While these methods improve average accuracy, they primarily focus on optimizing surface-level prompt formulations and rely on the LLM itself to internally resolve semantic ambiguity, often resulting in high variance under ambiguous or underspecified inputs.
Recent retrieval-augmented optimization approaches further attempt to refine prompt representations by reducing redundancy and improving retrieval alignment, yet still depend on implicit disambiguation within the model~\cite{guo2025hash,ou2025accelerating}.

\pb{Semantic Ambiguity Resolution.}
Semantic ambiguity has long been studied in natural language processing, particularly in tasks such as word sense ambiguation, co-reference resolution, and pragmatic inference~\cite{navigli2009word, hobbs1978resolving, kehler2002coherence}.
More recently, ambiguity-aware modeling has been explored in neural settings, including uncertainty-aware parsing~\cite{duan2024shifting, liu2025uncertainty}, contrastive disambiguation~\cite{arora2024contrastive, kibria2024functional}, and probabilistic semantic representations~\cite{min2020ambigqa}.
In the context of LLMs, ambiguity has been identified as a key source of reasoning errors and instability~\cite{dziri2022origin, ribeiro2020beyond}.
Some works attempt to mitigate ambiguity through paraphrasing, rephrasing, or clarification questions~\cite {si2022prompting,zhang2026learning}.
However, these approaches either defer ambiguity resolution to the LLM or require additional interaction rounds.
Moreover, they do not provide a structured mechanism for detecting, verifying, and resolving semantic conflicts prior to reasoning.
Research highlights the importance of explicitly modeling competing semantic hypotheses and validating their consistency, which can improve robustness under ambiguous inputs~\cite{guo2022beyond,guo2023learning}.

%% file: sections/method.tex
Our work introduces DisambiguSLM, a prompt optimization method that explicitly resolves semantic ambiguities in input prompts prior to LLM inference. 
We employ SLM as the core executor of DisambiguSLM to take advantage of its computational efficiency.


\subsection{Problem Formulation.}
We formulate LLM inference over an ambiguous input prompt as follows.
Given input query $q$, LLM $P_{LLM}$ aims to generate output text $y$ where $y \sim P_{LLM}(Y \mid Q)$.
$Q$ defines semantic representation of the query, and $Y$ represents semantic of the generated text.
Due to the expressiveness of users' input, $Q$ may contain ambiguities and logical conflicts.
For instance, \say{The boy saw the man with the telescope} may yields multiple interpretations, \textit{the man that uses a telescope is seen by the boy} or \textit{the boy is using a telescope to see the man}.
This may confuse the reasoning in LLM, and thus producing less satisfying answer.
This stems from high-entropy diffusion in the attention matrix $\textbf{A}_Q$ that leads to the expansion of the inference path $\mathcal{B}_Q$.
We define this input semantic risk as a measure of the inference uncertainty of the LLM.

\begin{equation}\label{eq:risk}
    \mathcal{R}(Q) = H(\textbf{A}_Q)+\gamma|\mathcal{B}_Q|
\end{equation}

\noindent
where $H(\textbf{A}_Q)$ is the attention entropy that characterizes the degree of dispersion in attention distribution, $|\mathcal{B}_Q|$ is the number of competing thinking branches, and $\gamma$ defines the coefficient that penalizes the branch expansion.
The semantic risk may include the following types:
\begin{itemize}
    \item \textit{Referential ambiguity} refers to the presence of unclear referents.
    \item \textit{Missing or implicit assumption} refers to the omission of key premises or background assumptions in the input.
    \item \textit{Temporal ambiguity} refers to the unclear temporal relationship or sequence of events.
\end{itemize}
Our work aims to minimize the semantic risk by transforming the semantics of the input prompt into a disambiguous prompt representation $Q'$.

\begin{equation}
    Q' = \Phi(Q), \quad \text{with} \quad Q' = \arg\min_{Q} \mathcal{R}(Q)
\end{equation}

\begin{figure*}
    \centering
    \includegraphics[width=.9\linewidth]{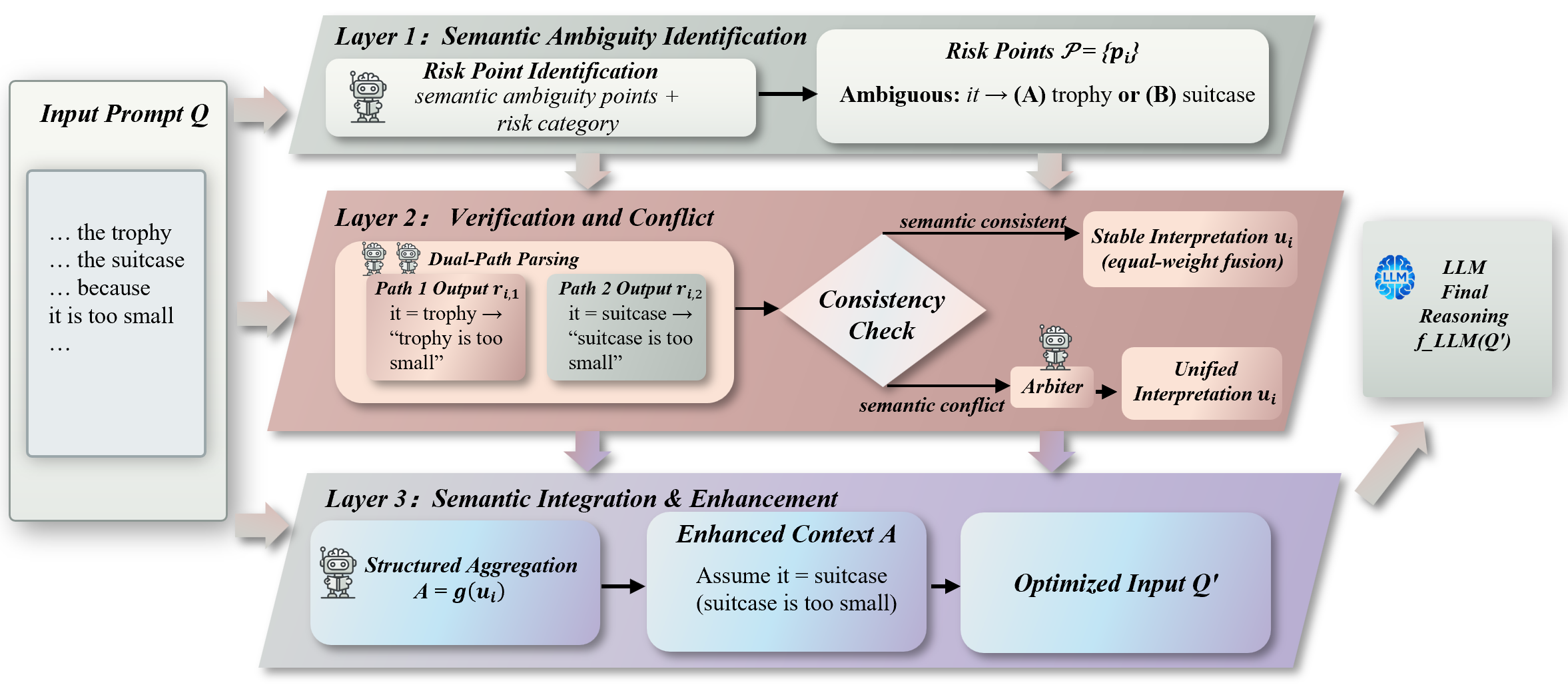}
    \caption{Framework of DisambiguSLM.
    We propose a prompt optimization method via semantic disambiguation. Specifically, we identify semantic risks, perform dual-path verification and conflict resolution, and conduct semantic integration to produce a non-ambiguous input prompt for LLM inference.}
    \label{fig:framework}
\end{figure*}

\subsection{Technical Framework}
We aim to transform the input prompt with the assistance of an SLM to identify and resolve semantic risk in the input prompt.
Figure~\ref{fig:framework} illustrates our method.

\pb{Layer 1: Semantic Risk Identification.}
We employ SLM as a global scanner to perform rigorous semantic scanning on $Q$ to locate abnormal regions that may cause instability in LLM inference.
SLM receives an input query $q$ and produces a set of risk points $\mathcal{P}$.

\begin{equation}
    \{\textbf{p}_i\}_{i=1}^n \in \mathcal{P}, \quad where \quad \textbf{p}_i = \left[s_i, t_i, \tau_i\right]^T 
\end{equation}

\noindent
Elements $s_i$, $t_i$, $\tau_i$ define 2D location information and risk type (e.g., ambiguity, logical gap).
We restrict the SLM role strictly for risk positioning and category identification to keep the computing budget low.

\pb{Layer 2: Consistency Verification and Conflict Resolution.}
After identifying the risk point, we process it in the next step, which is verification and resolution.
We pass each risk point $\textbf{p}_i$ to two independent SLMs that generate two semantic interpretations $r_{i,1}$ and $r_{i,2}$.
We then check the consistency $c_i$ of these two interpretations by calculating their semantic embedding similarity as follows:
\begin{equation}
    c_i = \text{cosine sim}(h(r_{i,1}), h(r_{i,2}))
\end{equation}

\noindent
The two interpretations are defined as \emph{consistent} when $c_i \geq \delta$, and \emph{inconsistent} otherwise.
For a consistent pair, we perform equal-weight fusion by \emph{concatenating} the two interpretations, without assigning preference to either one, to form a stable explanation $u_i$.
This concatenated representation explicitly preserves complementary semantic information and is subsequently normalized in the semantic integration stage.
Otherwise, when the two interpretations are inconsistent, we feed both interpretations to the SLM, together with the original problem context, to produce a self-consistent unified explanation $u_i$ that resolves the semantic conflict.

\pb{Layer 3: Semantic Integration and Enhancement.}
After verification, we feed the resolved risk point $u_i$ to another SLM to perform structured logic enhancement.
Specifically, we organize all the resolved representations into the SLM to normalize the information and generate a new enhanced context $A$.

\begin{equation}
    A = \mathrm{Aggregate}\Big( \{ \mathrm{SLM}(u_i) \}_{i=1}^n \Big)
\end{equation}

\noindent
We utilize this new representation, together with the original query $Q$, as a semantically unambiguous input $Q'$ to the LLM.

%% file: sections/theory.tex
We conjecture that unstable LLM reasoning is due to the presence of semantic risk in the input query $q$.
These risk points increase the entropy of the internal attention mechanism, thereby expanding the reasoning branch.
To verify this hypothesis, we conduct an observation on the evolution of the model's attention mechanism $\textbf{A}_{Q'}$ and analyze the contribution of DisambiguSLM as a stabilizer.
Particularly, we perform a comparative analysis of the impact of original query $Q$ and unambiguous query $Q'$ on Llama-3-8b.
We utilize the following demo prompt for analysis.

\small
\begin{Verbatim}[breaklines=true, commandchars=\\\{\}, mathescape=true]
Problem: 
Company A starts with \$1000. 
Company B starts with \$2000. 
There are two potential costs: 
Marketing (\$400) and R\&D (\$300).
Company A pays for Marketing. 
It then transfers half of the remainder to Company B.
Later, Company B invests 20\% of it into R\&D.
Question: 
How much money does Company B have left?
Let's think step by step.
\end{Verbatim}

\normalsize
\noindent
The pronoun \say{it} in the \textit{Company B invests 20\% of it} is semantically ambiguous as it may refer to the holding amount of B (\$2000) or to the amount received from A (\$300).
Thus, with this demo prompt, we analyze the attention distribution from \say{it} token to all preceding tokens in the prompt.

\begin{figure}[!htb]
    \centering
    \includegraphics[width=\linewidth]{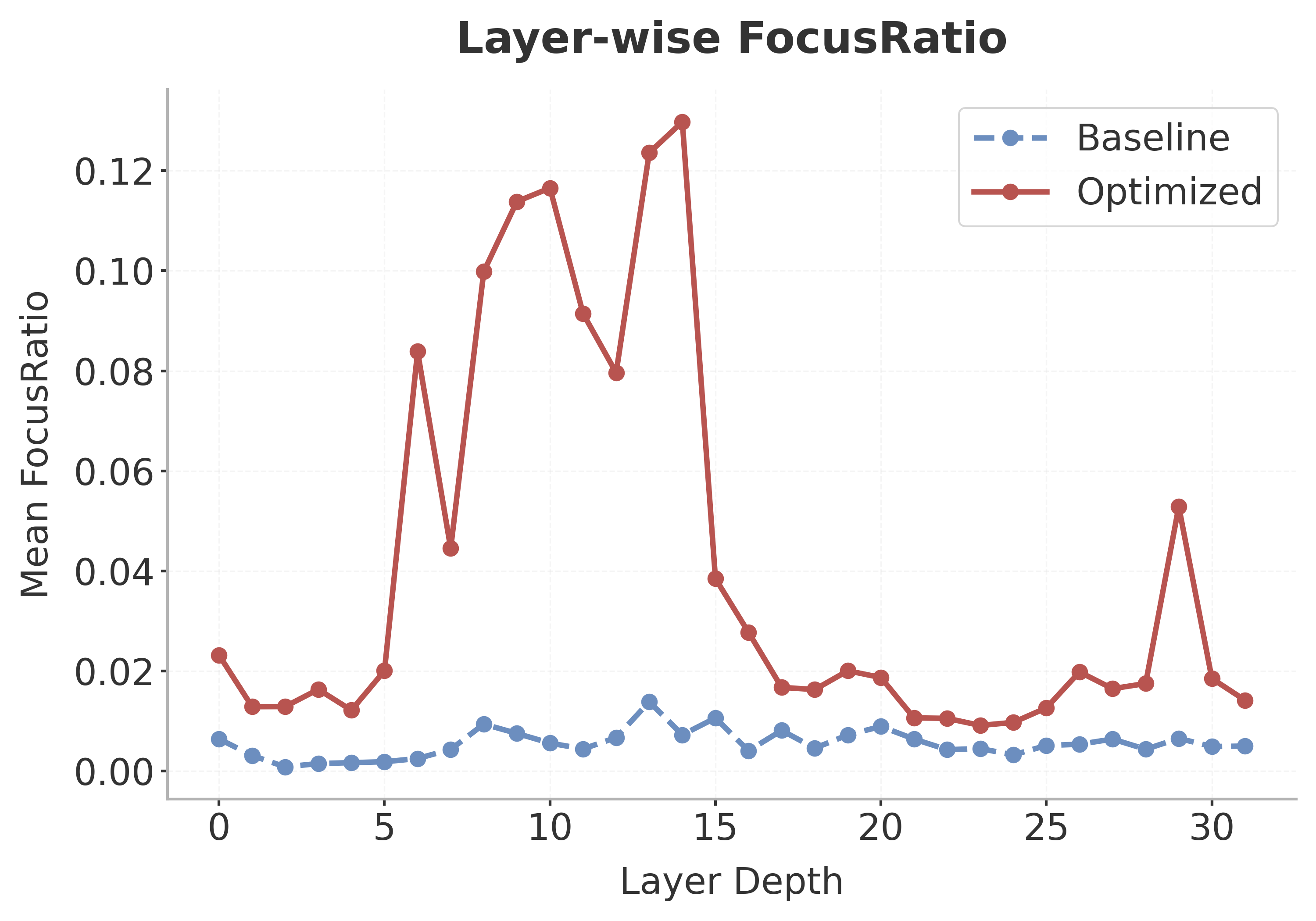}
    \caption{Layer-wise focus ratio comparison between $Q$ and $Q'$. $Q'$ significantly improves the attention focus in reasoning layers.}
    \label{fig:focusratio}
\end{figure}

\pb{Layer-wise Attention Focus.}
We measure the layer-wise attention focus ratio of the \say{it} token to all preceding tokens to understand how it evolves.
We define focus ratio (FR) as follows.

\begin{equation}
    \text{FocusRatio}(q, T) = \sum_{i \in T} \textbf{A}_{q,i}
\end{equation}


\noindent
where $T$ refers to the subset of preceding token indices of the semantically correct references (e.g., \say{remainder} token) and $\textbf{A}_q$ defines the normalized attention weight.
We plot layer-wise focus ratio by taking the mean of the attention score over queries.
Figure~\ref{fig:focusratio} shows that the optimized representation $Q'$ exhibits a significantly higher focus ratio than the original representation $Q$ across all layers.

\noindent
Focus ratio of $Q'$ exhibits immense growth in the early layer (e.g., 8 to 15) with a maximum increase of approximately eight to ten times.
Meanwhile, the focus ratio of $Q$ is relatively constant focus across all layers.
This phenomenon potentially demonstrates that upstream prompt disambiguation suppresses the ineffective diffusion of inference branches.
Instead of competing against diverse reasoning branches, the model precisely allocates computational resources to the key semantic anchor resolved by SLM.

\begin{figure}[!htb]
    \centering
    \includegraphics[width=\linewidth]{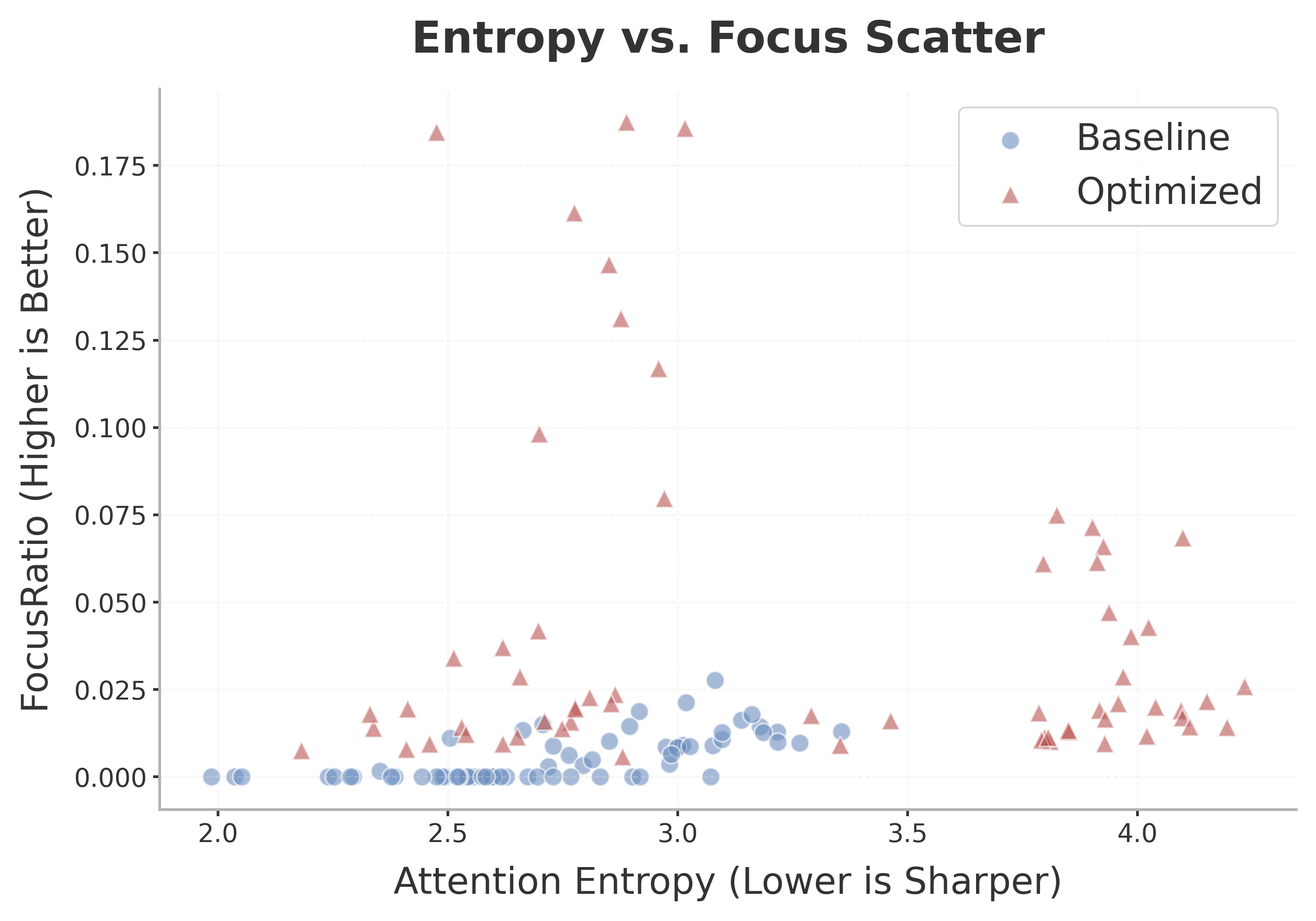}
    \caption{Comparison of joint distribution of entropy-focus ratio between $Q$ and $Q'$. $Q'$ shifts high focus region for low entropy and low focus region for high entropy.}
    \label{fig:entropyfocus}
\end{figure}

\pb{Entropy-Focus Distribution.}
Besides FocusRatio, we calculate Shannon entropy of the normalized attention weight as follows.

\begin{equation}
H(q)=-\sum_{i=1}^n A_{q,i}\log(A_{q,i}+\epsilon)
\end{equation}

\noindent
where $\epsilon$ is small constant for numerical stability.
We plot the distribution of entropy and focus ratio to examine the effect of query disambiguation on the information theory level.
Figure~\ref{fig:entropyfocus} shows that the distribution of $Q$ is concentrated in the low focus region.
Meanwhile, the distribution of $Q'$ significantly shifts to the high focus region for low entropy and low focus region for high entropy.
This phenomenon demonstrates the effectiveness of the enhanced context $A$ over the original query $Q$.
The intervention of SLM not only enables the model to identify key semantic anchors accurately but also reduces uncertainty during inference by eliminating semantic noise.

\begin{figure}[!htb]
    \centering
    \includegraphics[width=\linewidth]{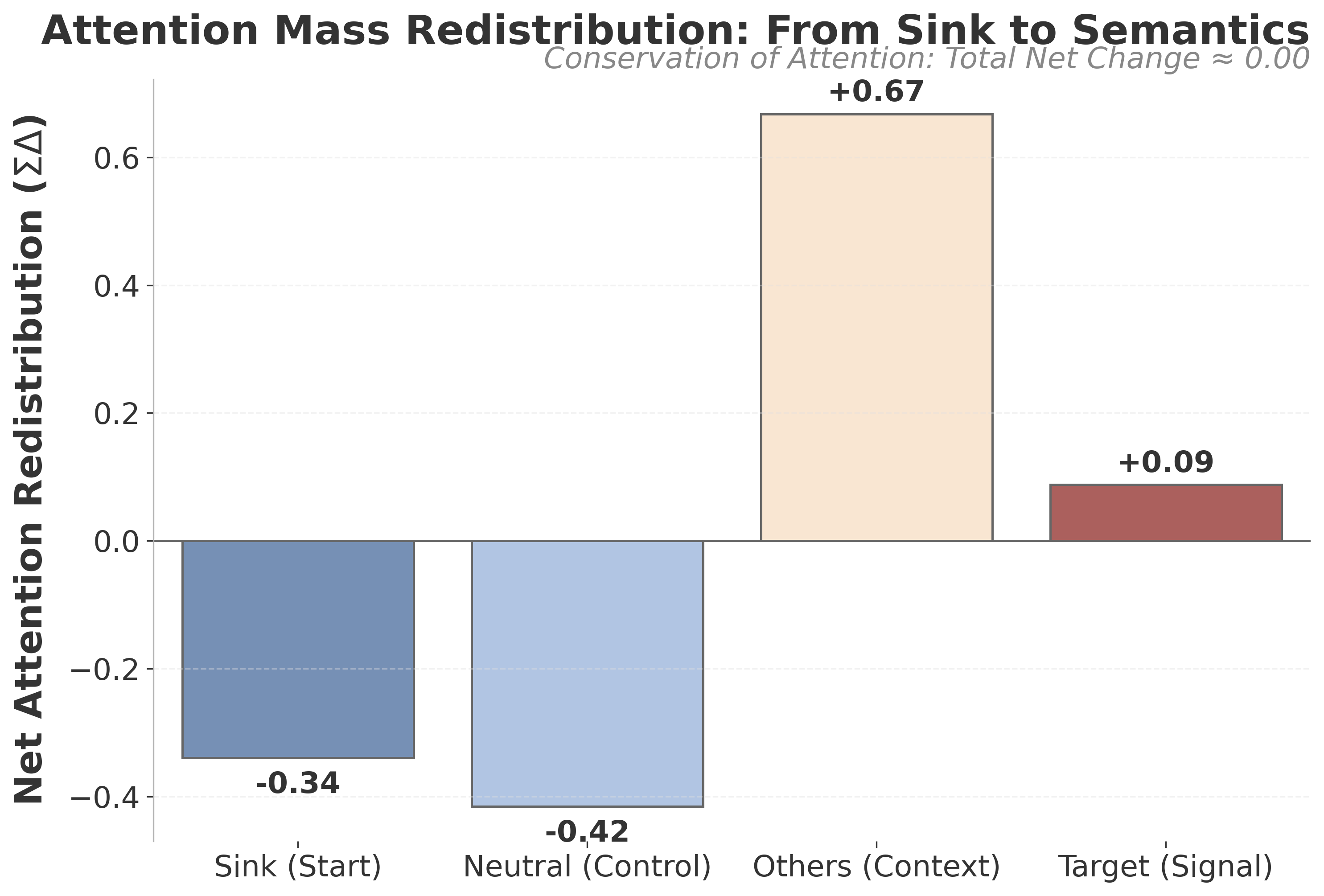}
    \caption{Token-wise attention reallocation from $Q$ to $Q'$. $Q'$ reallocate attention from sink tokens and stopwords to semantically-meaningful tokens.}
    \label{fig:redistribution}
\end{figure}

\pb{Token-Wise Attention Redistribution.}
To understand how $Q'$ changes the attention allocation in $Q$ at a fine-grained level, we plot the attention reallocation on each token type.
We categorize tokens into four types: sink token, neutral tokens, target tokens, and others.
Sink token includes start token \say{$<$|begin\_of\_text|$>$}, which typically behaves as an attention sink~\cite{xiao2023efficient}.
Neutral tokens include stop-word tokens (e.g., \textit{of}, \textit{is}).
Target token is the expected anchor tokens needed to answer the question correctly (e.g., \say{\$300}).
Others include words outside of the above categories but required by the target model to infer the information for answering the question.
Figure~\ref{fig:redistribution} demonstrates that $Q'$ withdraws a considerable amount of attention from sink token and neutral tokens and reallocates those to target and other tokens, which are more semantically meaningful in answering the question.

%% file: sections/results.tex
This section elaborates on the evaluation of our DisambiguSLM method.
We utilize Llama-3.2-1B as SLM and GPT-4o-mini as LLM.
This small-sized language model is sufficient to function as a semantic controller for all related subtasks in our method.
Unless stated otherwise, we set the hyperparameters as follows: similarity threshold to $\delta = 0.8$, temperature to $0.2$, and random seeds in the range 2025–2029. 
We utilize five samples per input.

\begin{table*}[!htb]
\small
    \centering
    \begin{threeparttable}
    \begin{tabular}{|p{40pt}|c|c|c|c|c|c|c|c|}
    \hline
        \textbf{Method} & \textbf{LLM} & \textbf{GPQA} & \textbf{AGIEval-MATH} & \textbf{LIAR} & \textbf{WSC} & \textbf{BBH-Navigate} & \textbf{Avg. Perf.} & \textbf{Avg. Cost (\$)} \\
        \hline
        \multirow{3}{50pt}{Na\"ive prompting} & GPT & 38.9 & 42.1 & 63.5 & 72.4 & 91.3 & 61.6 & -\\
        & LLaMA & 35.8 & 39.6 & 59.9 & 68.4 & 87.1 & 58.2 & -\\
        & DeepSeek & 36.4 & 40.2 & 60.8 & 69.7 & 88.9 & 59.2 & - \\
        \hline
        \multirow{3}{40pt}{CoT} & GPT & 41.6 & 44.5 & 65.4 & 77.8 & 89.7 & 63.8 & - \\
        & LLaMA & 38.2 & 42 & 61.5 & 73.9 & 86.3 & 60.4 & - \\
        & DeepSeek & 38.7 & 42.6 & 62.1 & 74.9 & 87.5 & 61.2 & - \\
        \hline
        \multirow{3}{40pt}{Rephrase} & GPT & 40.2 & 42.1 & 50.5 & 79.1 & 93.5 & 61.1 & -\\
        & LLaMA & 37.1 & 39.8 & 48.6 & 74.2 & 89.6 & 57.93 & -\\
        & DeepSeek & 37.9 & 40.3 & 49.2 & 75.6 & 90.8 & 58.8 & -\\
        \hline
        \multirow{3}{40pt}{Step-back} & GPT & 42.4 & 47.5 & 62.8 & 78.7 & 93.5 & 65 & -\\
        & LLaMA & 39.1 & 44.1 & 59.7 & 74 & 90.2 & 61.4 & -\\
        & DeepSeek & 39.8 & 45.2 & 60.4 & 75.1 & 91.2 & 62.3 & -\\
        \hline
        \multirow{3}{40pt}{APE} & GPT & 41.1 & 44.4 & 65.9 & 80.2 & 92.5 & 64.8 & 9.07\\
        & LLaMA & 38 & 42.5 & 62.6 & 76.1 & 89.3 & 61.7 & 9.07\\
        & DeepSeek & 38.5 & 42.9 & 63.4 & 77.3 & 90.1 & 62.4 & 9.07\\
        \hline
        \multirow{3}{40pt}{OPRO} & GPT & 43.3 & 46.1 & 67.6 & 80.2 & 95.8 & 66.6 & 4.51\\
        & LLaMA & 39.6 & 43.3 & 64.2 & 76.8 & 91.5 & 63.1 & 4.51\\
        & DeepSeek & 40.1 & 44 & 65.1 & 77.9 & 92.4 & 63.9 & 4.51\\
        \hline
        \multirow{3}{40pt}{Prompt
        Agent} & GPT & 41.3 & 41.4 & 64.1 & 82.7 & 95.7 & 65 & 2.71\\
        & LLaMA & 38.2 & 40.1 & 61.2 & 79.4 & 91.1 & 62 & 2.71\\
        & DeepSeek & 38.6 & 40.7 & 61.9 & 80.3 & 92.1 & 62.7 & 2.71\\
        \hline
        \multirow{3}{40pt}{Prompt Breeder} & GPT & 40.9 & 45.9 & 63.2 & 76.7 & 96.3 & 64.5 & 4.82\\
        & LLaMA & 37.6 & 42.9 & 60.3 & 74.1 & 92 & 61.4 & 4.82\\
        & DeepSeek & 38.1 & 43.8 & 61.1 & 74.8 & 93 & 62.2 & 4.82\\
        \hline
        \multirow{3}{40pt}{TextGrad} & GPT & 40.2 & 44.4 & 65.7 & 78 & 91.3 & 63.9 & 13.14\\
        & LLaMA & 36.9 & 41.8 & 62 & 75 & 88.7 & 60.9 & 13.14\\
        & DeepSeek & 37.5 & 42.1 & 63 & 75.9 & 89.4 & 61.6 & 13.14\\
        \hline
        \multirow{3}{40pt}{SPO} & GPT & 43.6 & 46.1 & 67.1 & 82 & 97.2 & 66.9 & 0.15\\
        & LLaMA & 40 & 43.5 & 63.8 & 79.6 & 93.1 & 64 & 0.15\\
        & DeepSeek & 40.5 & 44.2 & 64.7 & 80.1 & 94.3 & 64.8 & 0.15\\
        \hline
        \multirow{3}{40pt}{\textbf{Disambigu SLM (Ours)}} & GPT & \textbf{44} & \textbf{46.7} & \textbf{69} & \textbf{85.2} & \textbf{98} & \textbf{68.6} & \textbf{0.02}\\
        & LLaMA & \textbf{41.1} & \textbf{44.2} & \textbf{65.9} & \textbf{83.8} & \textbf{94.8} & \textbf{66} & \textbf{0.02}\\
        & DeepSeek & \textbf{41.6} & \textbf{44.9} & \textbf{66.6} & \textbf{83} & \textbf{95.4} & \textbf{66.3} & \textbf{0.02}\\
        \hline
    \end{tabular}
    \caption{Performance of prompting techniques across various reasoning benchmarks.}
    \label{tab:main}
    \end{threeparttable}
\end{table*}

\subsection{Main Result}
We evaluate the performance of our method on three LLMs: GPT-4o-mini, LLaMa-3-70B, and DeepSeek-V3 to cover diverse architectures and deployment settings, enabling a fair evaluation of robustness and cross-model generalization.
We compare our method against several strong baselines, which include direct prompting, chain-of-thought (CoT)~\cite{wei2022chain}, Rephrase~\cite{deng2023rephrase}, Step-back~\cite{zheng2023take}, APE~\cite{zhou2022large}, OPRO~\cite{yang2023large}, PromptAgent~\cite{wang2023promptagent}, PromptBreeder~\cite{fernando2023promptbreeder}, TextGrad~\cite{yuksekgonul2024textgrad}, and SPO~\cite{xiang2025self}.
We evaluate on five widely used reasoning benchmarks: GPQA~\cite{rein2024gpqa}, AGIEval-MATH~\cite{zhong2024agieval}, LIAR~\cite{wang2017liar}, WSC~\cite{levesque2012winograd}, and BBH-Navigate~\cite{suzgun2023challenging}.
These benchmarks encompass a diverse range of reasoning skills, including factual reasoning, mathematical problem solving, logical consistency checking, reference resolution, and multi-step navigation reasoning. 
Particularly, WSC and LIAR are known to be highly sensitive to semantic ambiguity and underspecified inputs, making them suitable for evaluating robustness under semantic risk.
We average the accuracy over three independent runs.
Additionally, we measure \textit{Avg. Cost} as the cost required for making an inference call to the optimizer model to obtain the final prompt.
Note that this cost excludes execution-time inference cost from calling the target LLM. 
Table~\ref{tab:main} presents our results.
Across all LLMs, our method consistently outperforms conventional prompting and prior prompt optimization, with large gains on ambiguity-sensitive tasks such as WSC and BBH-Navigate. 
By resolving semantic uncertainty at the input level, it stabilizes attention and reasoning trajectories before LLM execution. 
Gains are largest in LLaMA-3-70B, indicating that models are more sensitive to ambiguity and benefit most from explicit semantic rectification. 
Overall, input-level semantic risk mitigation with SLM provides an efficient low-cost path to improving reasoning robustness beyond surface-level prompt optimization.

\begin{table*}[!htb]
\small
    \centering
    \begin{threeparttable}
    \begin{tabular}{|c|c|c|c|c|c|c|c|c|c|}
    \hline
        \multirow{2}*{\textbf{Method}} & \multicolumn{3}{c|}{\textbf{Acc@1 (\%)}} & \multicolumn{3}{c|}{\textbf{Majority Acc (\%)}} & \multicolumn{3}{c|}{\textbf{Disagreement Rate (\%) $\downarrow$}}\\
        \cline{2-10}
        & \textbf{GPQA} & \textbf{LIAR} & \textbf{WSC} & \textbf{GPQA} & \textbf{LIAR} & \textbf{WSC} & \textbf{GPQA} & \textbf{LIAR} & \textbf{WSC} \\
        \hline
        Na\"ive prompting & 38.9 & 63.5 & 72.4 & 40.2 & 65 & 74.1 & 24.7 & 26.9 & 21.5\\
        CoT & 41.6 & 65.4 & 77.8 & 43.9 & 67.2 & 80.4 & 22.1 & 24.6 & 18.6\\
        Rephrase & 40.2 & 50.5 & 79.1 & 42.1 & 52.8 & 81 & 23.4 & 29.4 & 17.9\\
        Step-back & 42.4 & 62.8 & 78.7 & 45 & 65.1 & 81.3 & 20.8 & 25.3 & 17.1\\
        APE & 41.1 & 65.9 & 80.2 & 43.8 & 67.9 & 83.1 & 21.6 & 23.8 & 15.8\\
        OPRO & 43.3 & 67.6 & 80.2 & 46.5 & 70 & 83.6 & 18.9 & 21.4 & 15.2\\
        PromptAgent & 41.3 & 64.1 & 82.7 & 44.1 & 66.5 & 85.9 & 20.4 & 24.2 & 13.6\\
        PromptBreeder & 40.9 & 63.2 & 76.7 & 43 & 65.7 & 79.8 & 21.1 & 24.9 & 18.4\\
        TextGrad & 40.2 & 65.7 & 78 & 42.6 & 67.8 & 80.9 & 22.3 & 23.1 & 17.2\\
        SPO & 43.6 & 67.1 & 82 & 46.8 & 69.8 & 85.1 & 17.5 & 20.2 & 14.1\\
        \textbf{DisambiguSLM (Ours)} & \textbf{44} & \textbf{69} & \textbf{85.2} & \textbf{47.3} & \textbf{72.6} & \textbf{88.4} & \textbf{11.2} & \textbf{12.5} & \textbf{7.8}\\
        \hline
    \end{tabular}
    \caption{Stability and self-consistency comparison accross various prompting methods.}
    \label{tab:stability}
    \end{threeparttable}
\end{table*}

\pb{Stability and Consistency.}
Further, we measure the stability and self-consistency across prompt optimization methods.
Specifically, we utilize GPQA, LIAR, and WSC benchmarks.
We report single-sample accuracy (Acc@1), majority-vote accuracy, and disagreement rate.
Disagreement rate measures the proportion of inputs whose sampled outputs are not identical.
Lower disagreement rates indicate higher output stability.
Table~\ref{tab:stability} demonstrates that across all datasets, prompt optimization methods exhibit high sampling variance (15–30\% disagreement), as they rely on the target LLM to internally resolve semantic ambiguity. 
In contrast, our semantic risk rectification framework achieves the lowest disagreement rates on all datasets.
Particularly, LIAR and WSC indicate substantially more stable reasoning. 
These results show that input-level semantic rectification primarily improves consistency and reproducibility rather than merely boosting average performance.

\pb{Robustness on Ambiguity.}
Additionally, we evaluate the robustness of prompting methods on ambiguity-augmented benchmarks.
We randomly pick 200 samples from WSC, LIAR, and GPQA benchmarks.
We utilize GPT-4o-mini to generate a semantically ambiguous but still solvable variant by introducing underspecified references or mild logical uncertainty while preserving the original ground-truth answer.
Specifically, we utilize the following prompt.

\small
\begin{Verbatim}[breaklines=true, commandchars=\\\{\}, mathescape=true]
Instruction:
Rewrite the following question to make it slightly more semantically ambiguous, while keeping it logically solvable.
You may introduce ambiguity by:
$\bullet$ using unclear or underspecified references (e.g., pronouns or vague entities),
$\bullet$ omitting minor contextual details, or
$\bullet$ introducing mild logical uncertainty that requires inference.
Do NOT change the underlying correct answer.
The rewritten question should remain natural, fluent, and answerable by careful reasoning.
Only output the rewritten question.
Original Question:
{Q}
\end{Verbatim}

\normalsize
\noindent
Table~\ref{tab:robust} shows that all methods degrade on ambiguity-augmented inputs, validating increased semantic uncertainty. 
Standard prompting (e.g., Na\"ive prompting, CoT, Rephrase) is especially brittle on WSC and LIAR, while prompt optimization methods yield only modest gains by relying on the target model to resolve ambiguity at inference. 
Our semantic risk rectification framework consistently achieves the best performance across all benchmarks, particularly on WSC and LIAR.
This demonstrates the effectiveness of input-level ambiguity resolution. 
Stable performance on GPQA further confirms robustness without sacrificing knowledge-intensive reasoning.

\begin{table}[!htb]
\small
    \centering
    \begin{threeparttable}
    \begin{tabular}{|m{70pt}|c|c|c|c|}
    \hline
        \textbf{Method} & \textbf{WSC} & \textbf{LIAR} & \textbf{GPQA} & \textbf{Avg.}\\
        \hline
        Na\"ive prompting & 66.8 & 58.9 & 36.5 & 54.1\\
        CoT & 71.5 & 61.7 & 38.9 & 57.4\\
        Rephrase & 72.8 & 47.6 & 37.1 & 52.5\\
        Step-back & 73.1 & 60.8 & 39.7 & 57.9\\
        APE & 74.6 & 62.4 & 38.8 & 58.6\\
        OPRO & 75.2 & 64.1 & 40.3 & 59.9\\
        PromptAgent & 77.4 & 62 & 39.2 & 59.5\\
        PromptBreeder & 70.9 & 60.1 & 38.5 & 56.5\\
        TextGrad & 72 & 62.7 & 38.9 & 57.9\\
        SPO & 77.9 & 64.8 & 41.1 & 61.3\\
        \textbf{DisambiguSLM (Ours)} & \textbf{82.6} & \textbf{68.9} & \textbf{42.8} & \textbf{64.8}\\
        \hline
    \end{tabular}
    \caption{Robustness analysis on ambiguous prompt accross various prompting methods.}
    \label{tab:robust}
    \end{threeparttable}
\end{table}

\subsection{Ablation Studies}

\pb{Modular Contributions.}
We analyze the contribution of each element in our method.
Specifically, we evaluate the following scenarios: eliminating layers 2 and 3 (w/o L2 \& L3), changing dual to single verification on layer 2 (dual $\xrightarrow{}$ single), eliminating conflict resolution mechanism on layer 2 (w/o CR-L2), and eliminating only layer 3 (w/o L3).
We perform this ablation on three benchmarks, GPQA, LIAR, and WSC.
The results in Table~\ref{tab:module}, demonstrate that all modules are essential.
Removing dual-path verification causes major drops on LIAR and WSC, underscoring the need for redundant interpretation and consistency checks. 
Disabling conflict resolution further degrades performance, confirming the importance of explicitly resolving contradictory semantics. 
Variants that only detect risk without disambiguation perform worst, especially on WSC.
This demonstrates that gains come from active semantic rectification, not mere risk awareness. 
Removing structured aggregation on stage 3 also hurts performance, highlighting the value of a unified semantic representation for robust reasoning.

\begin{table}[!htb]
\small
    \centering
    \begin{threeparttable}
    \begin{tabular}{|c|c|c|c|c|}
    \hline
        \textbf{Module} & \textbf{GPQA} & \textbf{LIAR} & \textbf{WSC} & \textbf{Avg.}\\
        \hline
        Baseline (Ours) & 44 & 69 & 85.2 & 66.1\\
        \hline
        w/o L2 \& L3 & 40.9 & 64.8 & 76.5 & 60.7\\
        dual $\xrightarrow{}$ single & 42.7 & 66.2 & 81.6 & 63.5\\
        w/o CR-L2 & 43.2 & 66.9 & 82.3 & 64.1\\
        w/o L3 & 43.4 & 67.5 & 83.4 & 64.8\\
        \hline
    \end{tabular}
    \caption{Robustness analysis on ambiguous prompt across various prompting methods.}
    \label{tab:module}
    \end{threeparttable}
\end{table}

\pb{Sensitivity on SLM Variants.}
We examine the effect of SLM size and type on the prompting performance.
We utilize three benchmarks for this evaluation: GPQA, LIAR, and WSC.
Table~\ref{tab:slm} shows that a lightweight 1B model captures most benefits of semantic risk rectification. 
Scaling to 1.5B yields only marginal gains.
This shows that the framework’s effectiveness stems from structured semantic rectification rather than model scale, and that small models are sufficient for efficient semantic preprocessing.

\begin{table}[!htb]
\small
    \centering
    \begin{threeparttable}
    \begin{tabular}{|c|c|c|c|c|}
    \hline
        \textbf{SLM} & \textbf{GPQA} & \textbf{LIAR} & \textbf{WSC} & \textbf{Avg.}\\
        \hline
        Baseline (Ours) & 44 & 69 & 85.2 & 66.1\\
        \hline
        Qwen-2.5-1.5B & 44.2 & 69.3 & 85.7 & 66.4\\
        GPT-3.5-turbo & 44.7 & 70.1 & 86.6 & 67.1\\
        GPT-4o & 45.3 & 70.8 & 87.4 & 67.8\\
        \hline
    \end{tabular}
    \caption{The effect of SLM type on the reasoning performance.}
    \label{tab:slm}
    \end{threeparttable}
\end{table}

\begin{figure}[!htb]
    \centering
    \includegraphics[width=\linewidth]{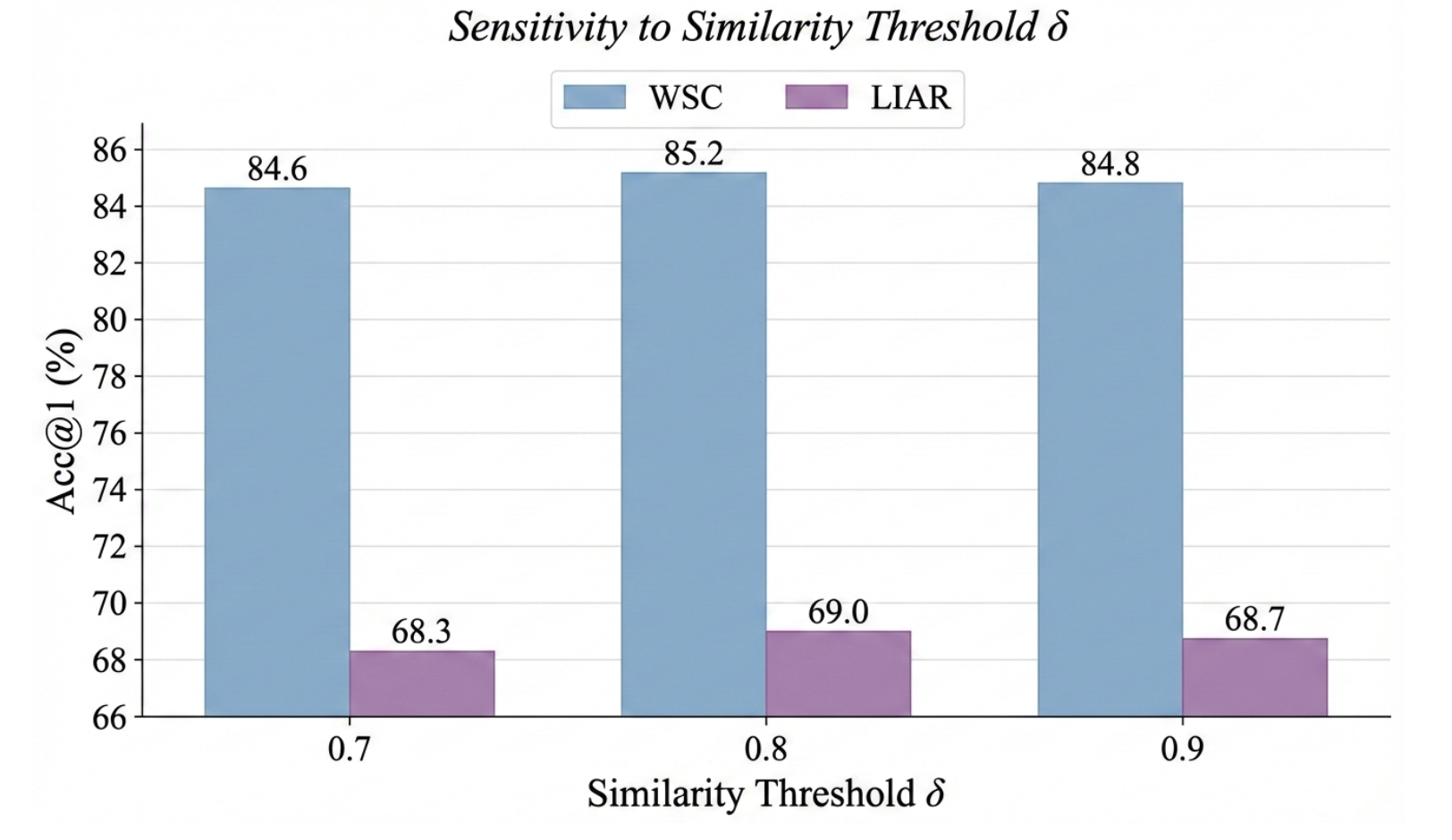}
    \caption{Evaluation on the sensitivity of similarity threshold.}
    \label{fig:threshold}
\end{figure}

\pb{Sensitivity of Hyperparameter.}
To verify the robustness of DisambiguSLM to the selection of similarity threshold $\delta$, we conducted parameter sensitivity experiments on different threshold settings for two benchmark tasks (WSC and LIAR) that are most sensitive to semantic ambiguity. 
We perform this experiment under a similar setup to the main experiment.
Figure~\ref{fig:threshold} demonstrates that within a reasonable threshold range, the model performance varies only slightly, and the overall trend remains consistent. 
This result indicates that DisambiguSLM is not highly sensitive to the similarity threshold, and its performance improvement mainly comes from the structured semantic risk identification and resolution mechanism, rather than fine-tuning of a specific threshold.

%% file: sections/conclusion.tex
We propose a prompt optimization method pre-LLM inference where SLM acts as a cheap yet well-performing evaluator.
The core idea of our method is the resolution of semantic ambiguities or risks in the original prompt through the assistance of SLM.
Our method demonstrates effectiveness in a wide range of reasoning benchmarks
against state-of-the-art prompting methods.
Our work expands the exploration of SLM as an assistant model for LLM in resolving the semantic confusion of user input.

%% file: sections/discussion.tex
We present DisambiguSLM as prompt optimization method via ambiguity resolution that leverage SLM as the resolver for efficient execution.
DisambiguSLM aims to address semantic uncertainties at the input level, rather than correcting factual errors.
In fact-checking tasks, semantic disambiguation can only reduce ambiguity in understanding, but cannot replace external knowledge or evidence retrieval.
When task performance is primarily limited by the lack of facts rather than semantic ambiguity, the benefits of DisambiguSLM may be limited. 
For instance, given the prompt \say{The governor said the tax plan would help middle-class families}, DisambiguSLM will detect two semantic risks.
First, referential ambiguity in the specific referent of \textit{the tax plan}, which may correspond to different versions or different stages of policy plans.
Second, missing assumptions in the definition of \textit{help} (such as short-term tax cuts, long-term income growth, etc.) and a criterion for defining \textit{middle-class families}.
During the processing of DisambiguSLM, SLM can identify the above semantic risks and generate several reasonable semantic clarification explanations. 
However, these explanations are all based on the factual premise of the input statement itself.

%% file: sections/appendix.tex
\subsection{Analytical Assumption}

\pb{Conceptual Role of Semantic Risk Function.}
We introduces semantic risk function (Equation~\ref{eq:risk}) to characterize how semantic ambiguity in input prompts leads to instability in the inference process of LLMs. 
This function combines the entropy of the attention distribution and the number of potentially competing inference branches to explain the impact of semantic ambiguity on the diffusion of inference paths at the conceptual level. 
Note that this semantic risk function is only used for conceptual modeling, aiming to help understand the semantic ambiguity phenomena, rather than serving as an optimization objective or decision-making basis in the DisambiguSLM.
We do not compute the risk function at any stage of DisambiguSLM, and the weight parameters are not instantiated, adjusted, or used for any inference decision-making.
Further, the prompt repair process of DisambiguSLM does not depend on the numerical value or the minimized result of this function.

\subsection{Prompt Design}

\pb{Layer 1: Risk Identification.}
The goal of this stage is to identify key segments from the input prompt that may lead to semantic ambiguity or unstable reasoning without attempting to solve the problem.

\small
\begin{Verbatim}[breaklines=true, commandchars=\\\{\}, mathescape=true]
You are given a user question.
Your task is to identify spans or phrases that may introduce semantic ambiguity, underspecified references, or logical uncertainty that could affect reasoning stability.
Do NOT attempt to answer the question.
Do NOT resolve the ambiguity.
For each identified risk point, output:
(1) the exact text span, and
(2) a brief description of the type of semantic risk (referential ambiguity, missing assumption, temporal ambiguity).
User Question:
$\{\text{Input Prompt}\}$
\end{Verbatim}

\normalsize
\pb{Layer 2: Dual-Interpretation Generation.}
For each identified semantic risk point, SLM is independently invoked twice under the same context to generate two potential semantic interpretations, thereby covering different reasonable understandings.

\small
\begin{Verbatim}[breaklines=true, commandchars=\\\{\}, mathescape=true]
You are given a user question and a specific semantic risk point extracted from it.
Your task is to provide ONE plausible interpretation that clarifies the meaning of the risk point in the context of the question.
Focus only on explaining the risk point.
Do NOT solve the entire problem.
User Question:
$\{\text{Input Prompt}\}$
Semantic Risk Point:
$\{\text{Risk Span}\}$
\end{Verbatim}

\normalsize
\pb{Layer 2: Verification and Resolution.}
The system calculates the semantic similarity between two explanations and compares it with a preset threshold to produce the following conditions:
\begin{itemize}
    \item If two explanations are determined to be consistent, they will be equally weighted and fused through \textit{concatenation} to generate a stable explanation.
    \item Otherwise, proceed to the conflict resolution step.
\end{itemize} 

\small
\begin{Verbatim}[breaklines=true, commandchars=\\\{\}, mathescape=true]
You are given a user question and two different interpretations of the same semantic risk point.
Your task is to produce a single, self-consistent explanation that resolves the semantic conflict between the two interpretations, based on your explanation of the original question, and ensure logical consistency.
User Question:
$\{\text{Input Prompt}\}$
Interpretation 1:
$\{\text{Interpretation 1}\}$
Interpretation 2:
$\{\text{Interpretation 2}\}$
\end{Verbatim}

\normalsize
\pb{Layer 3: Semantic Integration and Enhancement.}
At this stage, SLM integrates all repaired semantic interpretations into a unified, structured enhanced semantic representation to assist subsequent LLM reasoning. 

\small
\begin{Verbatim}[breaklines=true, commandchars=\\\{\}, mathescape=true]
You are given a user question and a list of resolved semantic clarifications.
Your task is to integrate these clarifications into a concise and structured supplemental context that improves semantic clarity for downstream reasoning.
Do NOT answer the question.
Do NOT introduce new assumptions.
User Question:
$\{\text{Input Prompt}\}$
Resolved Clarifications:
$\{\text{Resolved Explanations}\}$
\end{Verbatim}